\documentclass[conference]{IEEEtran}
\IEEEoverridecommandlockouts

\usepackage{cite}
\usepackage{amsmath,amssymb,amsfonts}
\usepackage{algorithmic}
\usepackage{graphicx}
\usepackage{textcomp}
\usepackage{xcolor}
\def\BibTeX{{\rm B\kern-.05em{\sc i\kern-.025em b}\kern-.08em
    T\kern-.1667em\lower.7ex\hbox{E}\kern-.125emX}}
\usepackage{amsmath,graphicx}
\usepackage{booktabs}
\usepackage{multirow}
\usepackage{subfigure}
\usepackage{makecell}

\usepackage{amsmath}
\usepackage{color}
\usepackage{xcolor}
\usepackage{amsmath}
\usepackage{amssymb}

\newcommand{\KeyHeadword}[1]{\noindent\textbf{#1.}}

\usepackage{xcolor}
\usepackage{multirow}
\usepackage{bigstrut}
\usepackage{setspace}
\newcommand{\firstpara}[1]{\noindent\textbf{{#1}.}~}

\usepackage{array}
\usepackage{diagbox}
\usepackage{booktabs}
\usepackage{graphicx} 
\usepackage{subcaption}
\usepackage{enumitem}

\begin{document}
\title{Imperceptible  Adversarial  Attacks on  Point Clouds  Guided by  Point-to-Surface Field
}

\author{
\IEEEauthorblockN{\hspace{-0.2cm}Keke Tang\thanks{This work was supported in part by the National Natural Science Foundation of China (62472117, U2436208, 62406095), the Guangdong Basic and Applied Basic Research Foundation (2024A1515012064), and the Science and Technology Projects in Guangzhou (2025A03J0137).
}}
\IEEEauthorblockA{\hspace{-0.2cm}\textit{Guangzhou University} \\
\hspace{-0.2cm}tangbohutbh@gmail.com}

\and
\IEEEauthorblockN{Weiyao Ke}
\IEEEauthorblockA{\textit{Guangzhou University} \\
weiyaoke08@gmail.com}

\and
\IEEEauthorblockN{Weilong Peng\thanks{Keke Tang and Weiyao Ke contributed equally. Weilong Peng and Peican Zhu are joint corresponding authors.}}
\IEEEauthorblockA{\textit{Guangzhou University} \\
wlpeng@tju.edu.cn}

\and
\IEEEauthorblockN{\hspace{0.2cm}Xiaofei Wang\hspace{0.3cm}}
\IEEEauthorblockA{\hspace{0.2cm}\textit{SmartMore Corporation\hspace{0.3cm}} \\
\hspace{0.2cm}wxf9545@mail.ustc.edu.cn\hspace{0.3cm}}

\and
\IEEEauthorblockN{\hspace{0.5cm}Ziyong Du}
\IEEEauthorblockA{\hspace{0.5cm}\textit{Guangzhou University} \\
\hspace{0.5cm}duxiaoshuaicst@gmail.com}

\and
\IEEEauthorblockN{\hspace{0.8cm}Zhize Wu}
\IEEEauthorblockA{\textit{\hspace{0.8cm}Hefei University} \\
\hspace{0.8cm}wuzz@hfuu.edu.cn}

\and
\IEEEauthorblockN{\hspace{-0.2cm}Peizan Zhu}
\IEEEauthorblockA{\textit{\hspace{-0.2cm}Northwestern Polytechnical University} \\
\hspace{-0.2cm}ericcan@nwpu.edu.cn}

\and
\IEEEauthorblockN{\hspace{-0.2cm}Zhihong Tian}
\IEEEauthorblockA{\hspace{-0.2cm}\textit{Guangzhou University} \\
\hspace{-0.2cm}tianzhihong@gzhu.edu.cn}
}

\maketitle
\begin{abstract}

Adversarial attacks on point clouds are crucial for assessing and improving the adversarial robustness of 3D deep learning models. Traditional solutions strictly limit point displacement during attacks, making it challenging to balance imperceptibility with adversarial effectiveness. In this paper, we attribute the inadequate imperceptibility of adversarial attacks on point clouds to deviations from the underlying surface. 
To address this, we introduce a novel point-to-surface (P2S) field that adjusts adversarial perturbation directions by dragging points back to their original underlying surface. Specifically, we use a denoising network to learn the                             gradient field of the logarithmic density function encoding the shape's surface, and apply a distance-aware adjustment to perturbation directions during attacks, thereby enhancing imperceptibility. Extensive experiments show that adversarial attacks guided by our P2S field are  more imperceptible, outperforming state-of-the-art methods. 

\end{abstract}

\begin{IEEEkeywords}
 deep neural network, adversarial attacks,  imperceptibility, point clouds, surface
\end{IEEEkeywords}
\section{Introduction}
\label{sec:intro}

With the advancement of deep learning techniques~\cite{Lecun-2015-DeepLearning,Tang-DFN} and the increased availability of affordable depth-sensing devices, 3D point cloud perception using deep neural networks (DNNs) has become a prominent solution~\cite{guo2020deep,ioannidou2017deep,tang2022reppvconv}. However, recent studies have demonstrated that DNN classifiers are susceptible to adversarial attacks~\cite{Xiang-2019-3DAdversarialPCD,Liu-2019-extendingAdv3D}, where  imperceptible perturbations to input point clouds can result in incorrect predictions. This vulnerability poses significant challenges for their application in real-world scenarios. Therefore, investigating adversarial attacks on
3D 
DNN classifiers  is essential for evaluating and improving their adversarial robustness~\cite{tang2023enhancing,zhu2023improving,tang2024effective}.

To achieve imperceptible attacks on 3D point clouds, a classic approach is to employ constraints such as the $l_2$-norm, Chamfer distance, and Hausdorff distance to restrict point displacements~\cite{Xiang-2019-3DAdversarialPCD}. However, for adversarial attacks to be effective, these points must be displaced, making it challenging to balance imperceptibility and adversarial effectiveness. In practice, the displacement of points is not the primary cause of perceptibility; rather, it is the deviation of these points from the underlying surface of the point cloud that makes them noticeable~\cite{pistilli2020learning}. Therefore, as long as the points remain on the original surface, slightly larger displacements during attacks can still achieve imperceptibility.

In this paper, we introduce a novel point-to-surface (P2S) field for dragging perturbed points onto the surface during attacks, achieving enhanced imperceptibility. Specifically, we train a denoising network to learn the gradient field of the logarithmic density function encoding the shape surface. This field directs any point in Euclidean space toward the surface. Considering that points farther from the surface need more significant adjustments, we define a distance-aware P2S field magnitude. By iteratively dragging the initial perturbation direction onto the surface and then determining the perturbation magnitude, our adversarial attacks become more imperceptible. We validate the effectiveness of our solution by attacking various common 3D DNN classifiers. Extensive experimental results show that the generated adversarial point clouds are significantly more imperceptible than those produced by state-of-the-art methods.

Overall, our contribution is summarized as follows:

\begin{itemize}[itemsep=1pt,topsep=1pt,parsep=1pt]
\item We are the first to attribute the inadequate imperceptibility of adversarial attacks on 3D point clouds to the deviation from the underlying surface.

\item We devise a point-to-surface (P2S) field and a novel adversarial attack framework that employs this field to drag perturbed points onto the surface.

\item We show by experiments that adversarial attacks guided by the P2S field achieve superior performance in terms of imperceptibility.

\end{itemize}

\section{Related Work}
\label{sec:related}

\firstpara{Adversarial Attacks  on  Point Clouds}
Adversarial attacks~\cite{li2023hept,zhu2023sgma, ijcai2024p0063,yang2024c,fan2023DBA,He2025aaai,pan2024tcss}, aimed at generating samples that can mislead target networks, originated in 2D image classification and have been successfully extended to 3D point clouds.
Existing point cloud attacks are categorized into three types: addition-based, introducing independent points to induce errors~\cite{Xiang-2019-3DAdversarialPCD}; deletion-based, involving the removal of critical points to affect classification~\cite{Zheng-2019-PCDSaliency,Yang-2019-AdvAttackAndDefense,Wicker-2019-IterSaliencyOcc,zhang-2021-TopologyDestructionNetwork}; and perturbation-based, which involves altering existing points to facilitate attacks~\cite{Xiang-2019-3DAdversarialPCD,zhao2020isometry,Kim-2021-minimalAdv,tang2024symattack,Tang-2024-FLAT,tang2025EdgeAttack,Tang-2025-FLAT}. This paper  focuses on perturbation-based methods.

To achieve imperceptibility of attacks, a common approach is to apply constraints such as the $l_2$-norm, Chamfer distance, and Hausdorff distance between the original and adversarial point clouds~\cite{Xiang-2019-3DAdversarialPCD, Liu-2019-extendingAdv3D,Zhou-2020-LGGAN}. Beyond these standard constraints, GeoA$^3$~\cite{wen2020geometry} maintains local curvatures after the attack. More recent solutions guide perturbations along normal or tangential directions~\cite{liu2022imperceptible,huang2022shape,Tang-NTAttack}. In contrast, our approach achieves imperceptibility by dragging the perturbed points onto the surface.

\firstpara{Surface Modeling of Point Clouds}
Surface modeling of point clouds encompasses traditional geometric methods, such as Poisson surface reconstruction~\cite{kazhdan2006poisson}, implicit surface techniques like Signed Distance Functions~\cite{park2019deepsdf}, and recent deep learning approaches~\cite{cai2020learning}. In our work, we also employ deep learning to implicitly represent the surface. Our goal is to drag adversarially perturbed points back to the surface to achieve imperceptibility.

\section{Problem Formulation}
\label{sec:problem}

\firstpara{Preliminary on Adversarial Attacks}
Given a point cloud $P \in \mathbb{R}^{n \times 3}$ sampled from an object surface $S$ and its label $y \in \{1, \ldots, K\}$,  adversarial attack aims to mislead a 3D DNN classifier $f$ by feeding a perturbed point cloud $P^{'}$:
\begin{equation}
p{'}_i  = p_i +    \sigma_{p_i} \cdot \overrightarrow{d_{p_i}}, 
\label{eq:adv}
\end{equation}
where  $\sigma_{p_i}$ is the perturbation size for the  $i$-th point in $P$, i.e., $p_i$, and $ \overrightarrow{d_{p_i}}$ is the unit  perturbation direction.
Formally, this perturbation, $\sigma_{p_i} \cdot \overrightarrow{d_{p_i}}$,  can be computed by solving the following optimization problem, iteratively:
\begin{equation}
\setlength\abovedisplayskip{3pt}
\setlength\belowdisplayskip{3pt}
\min_{{\sigma_{p_i}}, {\overrightarrow{d_{p_i}}}} L_{mis} ({f}, P^{'}, y) + \lambda_1 C(P, P^{'}),
\end{equation}
where $L_{mis}(\cdot, \cdot, \cdot)$ is the misclassification loss (e.g., the negated cross-entropy loss), 
$P^{'}$ is the adversarial point clouds consists of $\{p{'}_i \}_{i=1:n}$, 
$C(\cdot, \cdot)$ is a  constraint to ensure imperceptibility, and $\lambda_1$ is a weighting parameter. In particular, our focus is on untargeted attacks, and targeted attacks can be similarly addressed.

\firstpara{Discussion}
Common choices for \(C(\cdot, \cdot)\), such as the $l_2$-norm, Chamfer distance, and Hausdorff distance, impose strict limits on point displacements, making it challenging to balance imperceptibility and adversarial effectiveness. In practice, if the points remain on the original surface $S$, slightly larger displacements can still achieve imperceptibility. Therefore, a feasible approach to achieving imperceptible adversarial attacks is to apply perturbations while dragging the points back towards the surface $S$.

\firstpara{Point-to-Surface Field-Guided Attacks}
Suppose there is a point-to-surface (P2S) field \(\mathcal{F}\) that, given a point \(q\), drags it closer to the surface \(S\), such that
\begin{equation}
D(q,S) > D(q + \mathcal{F}(q), S),
\end{equation}
where $D(\cdot,\cdot)$ measures the point-to-surface distance.

Therefore, adversarial points generated by P2S field-guided attacks can be formulated as:
\begin{equation}
p'_i \leftarrow p'_i + \mathcal{F}(p'_i).
\label{eq:adv_2}
\end{equation}
Compared to the original point, the updated point $p'_i$ is closer to the surface $S$.

\section{Method}

\begin{figure}
    \centering
    \includegraphics[width=0.91\linewidth]{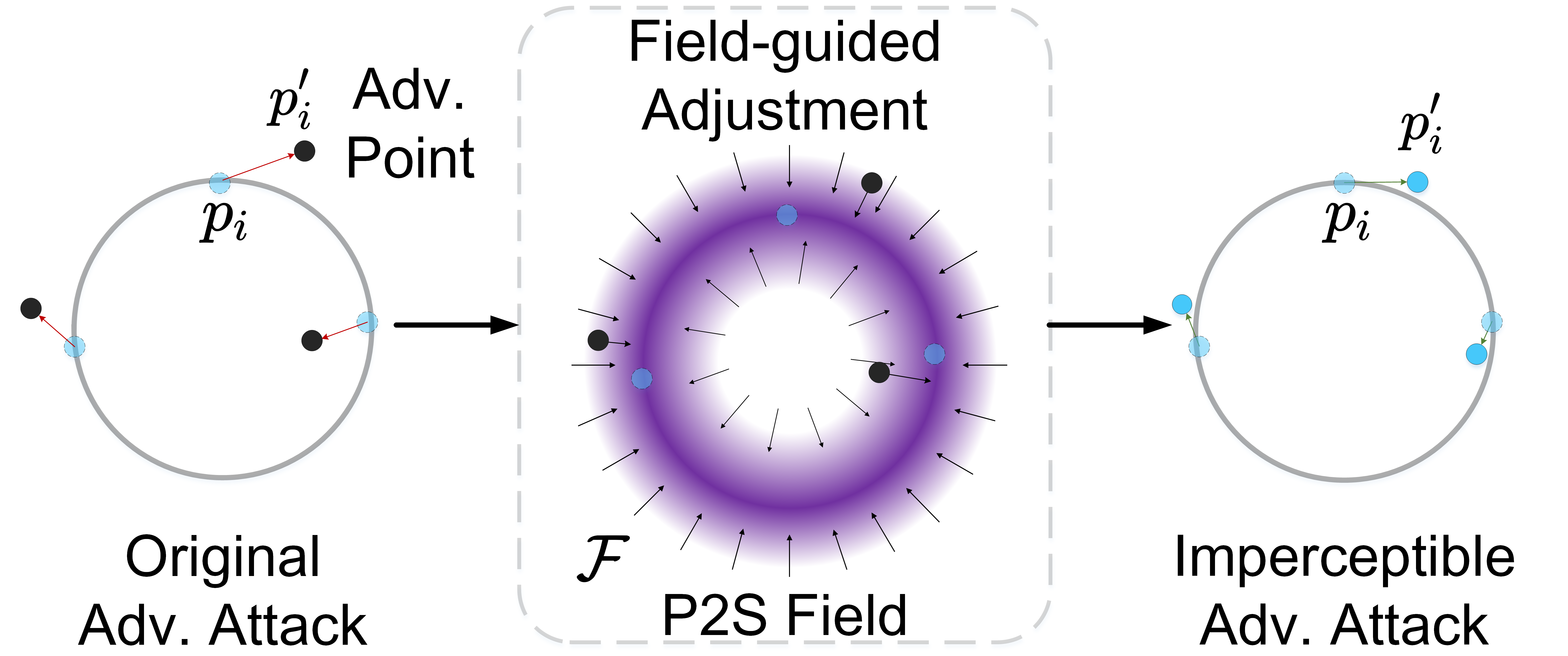}
    \vspace{-1mm}
    \caption{
Illustration of our point-to-surface (P2S) field-guided adversarial attacks. For each adversarial point in a point cloud, we adjust its direction using the P2S field to bring it closer to the surface, making the perturbation imperceptible.
    }
    \label{fig:method}
\end{figure}

In this section, we first introduce how to construct the point-to-surface (P2S) field using  DNN, and then outline our P2S field-guided adversarial attack framework. Please refer to Fig.~\ref{fig:method} for a demonstration.

\subsection{DNN-based Point-to-Surface Field}

We learn the point-to-surface (P2S) field using a DNN network. Specifically, we train the network to predict the gradient field from noisy point cloud data by minimizing the $l_2$ loss between the predicted gradients of the logarithmic density function encoding a shape and the ground-truth gradients estimated from the input point cloud, following~\cite{cai2020learning}.

Since the network has learned  how to move points towards high-density regions, i.e., the shape's surface $S$, 
the P2S field at the position \(q\) can be estimated as follows:
\begin{equation}
    \mathcal{F} (q) := \nabla_{q} \log Q_S(q),
\label{eq:gravityDir}
\end{equation}
where $Q_S(\cdot)$ approximates the true data distribution whose density concentrates near the surface $S$.

\subsection{P2S Field-guided Imperceptible Adversarial Attacks}

Given the clean point cloud $P$, we first randomly initializes the perturbation to
form the $0$-iteration adversarial point cloud $P'^{(0)}$, and then iteratively applying the following three steps.

\firstpara{Generating Initial Perturbations}
We employ IFGM~\cite{Dong-2020-FGM_PGD_IMPLEMENT} to generate initial perturbation directions for all points. It is noteworthy that alternative methods could also be employed to achieve similar effects.

\begin{table*}[!t]
\centering
\caption{
Comparison on the perturbation sizes required by different methods to reach their highest achievable ASR in the untargeted attack setting. The evaluation is conducted across different DNN classifiers on  ModelNet40 and ShapeNet Part.
}
\vspace{-1mm}
\label{tab:AttackSummaryTable}
\scalebox{0.8}{
\begin{tabular}{c|c|ccccccc|ccccccc}
\hline
\Xhline{1.0pt} 
\multirow{3}{*}{Model} & \multicolumn{1}{c|}{\multirow{3}{*}{Attack}} & \multicolumn{7}{c|}{ModelNet40} & \multicolumn{7}{c}{ShapeNet   Part} \\ \cline{3-16} 
 & \multicolumn{1}{c|}{} & ASR & CD & HD & $l_2$ & GR & Curv & EMD & ASR & CD & HD & $l_2$ & GR & Curv & EMD \\
 & \multicolumn{1}{c|}{} & (\%) & ($10^{-4}$) & ($10^{-2}$) &  &  & ($10^{-2}$) & ($10^{-2}$) & (\%) & ($10^{-4}$) & ($10^{-2}$) &  &  & ($10^{-2}$) & ($10^{-2}$) \\ \hline
\multirow{7}{*}{\rotatebox{90}{PointNet}} 
 & PGD & 100 & 7.155 & 5.025 & 0.981 & 0.302 & 1.624 & 2.315 & 100 & 13.172 & 17.068 & 1.569 & 0.521 & 3.679 & 3.358 \\
 & IFGM & 100 & 4.039 & 5.565 & 0.789 & 0.314 & 0.775 & 0.864 & 100 & 3.328 & 10.269 & 0.785 & 0.408 & 0.619 & 0.556 \\
 & GeoA$^3$ & 100 & 4.646 & 0.497 & 1.307 & 0.121 & 0.396 & 2.319 & 100 & 7.531 & 1.444 & 2.655 & 0.146 & 0.465 & 4.104 \\
 & 3d-Adv & 100 & 6.115 & 4.372 & 0.863 & 0.250 & 1.215 & 1.410 & 100 & 15.659 & 5.495 & 1.787 & 0.279 & 4.006 & 3.693 \\
 & SI-Adv & 100 & 2.768 & 2.595 & 0.731 & 0.220 & 0.271 & 0.725 & 100 & 3.435 & 3.692 & 0.881 & 0.233 & \textbf{0.441} & 0.825 \\
 & ITA & 100 & 2.747 & 0.414 & 0.534 & 0.122 & 0.555 & 1.214 & 100 & 5.872 & 1.917 & 1.002 & 0.181 & 1.016 & 2.035 \\
 & Ours & 100 & \textbf{1.110} & \textbf{0.358} & \textbf{0.366} & \textbf{0.115} & \textbf{0.260} & \textbf{0.449} & 100 & \textbf{2.822} & \textbf{1.347} & \textbf{0.780} & \textbf{0.145} & 0.492 & \textbf{0.543} \\\hline

\multirow{7}{*}{\rotatebox{90}{DGCNN}} 
 & PGD & 100 & 19.968 & 5.098 & 1.933 & 0.267 & 4.924 & 4.785 & 100 & 63.556 & 27.557 & 5.224 & 0.511 & 7.275 & 9.233 \\
 & IFGM & 100 & 15.791 & 12.391 & 1.622 & 0.363 & 2.849 & 3.777 & 100 & 19.623 & 26.040 & 2.069 & 0.504 & 4.954 & 4.387 \\
 & GeoA$^3$ & 100 & 7.566 & 0.546 & 1.585 & 0.119 & 0.741 & 3.083 & 100 & 27.612 & 3.748 & 5.798 & 0.199 & \textbf{1.695} & 7.502 \\
 & 3d-Adv & 100\ & 10.345 & 3.807 & 3.589 & 0.227 & 5.997 & 6.685 & 100 & 21.553 & 8.531 & 2.258 & 0.282 & 5.119 & 4.628 \\
 & SI-Adv & 100 & 7.146 & 1.691 & 1.087 & 0.143 & 0.666 & 2.495 & 100 & 11.685 & 3.019 & 1.772 & 0.160  & 2.054 & 3.646 \\
 & ITA & 100 & 3.249 & 0.524 & \textbf{0.552 }& 0.114 & 0.971 & 1.359 & 100 & 27.633 & 4.597 & 2.492 & 0.244 & 3.847 & 4.696 \\
 & Ours & 100 & \textbf{1.898} & \textbf{0.316} & 0.619 & \textbf{0.110} & \textbf{0.516} & \textbf{1.016} & 100 & \textbf{7.013} & \textbf{1.339} & \textbf{1.668} & \textbf{0.137} & 2.521 & \textbf{2.663} \\ \hline
 
\multirow{7}{*}{\rotatebox{90}{PointConv}} 
 & PGD & 100 & 14.551 & 2.216 & 1.442 & 0.184 & 3.491 & 3.862 & 100 & 42.202 & 9.949 & 3.784 & 0.252 & 6.866 & 7.277 \\
 & IFGM & 100 & 7.959 & 2.608 & 1.015 & 0.184 & 1.741 & 2.427 & 100 & 16.139 & 8.776 & 1.812 & 0.231 & 3.526 & 3.807 \\
 & GeoA$^3$ & 100 & 6.809 & 0.644 & 2.169 & 0.119 & 1.119 & 3.556 & 100 & 9.383 & 1.222 & 4.224 & 0.120 & \textbf{1.190} & 5.391 \\
 & 3d-Adv & 100 & 11.213 & 1.763 & 1.179 & 0.163 & 3.279 & 2.807 & 100 & 21.034 & 3.687 & 2.277 & 0.193 & 4.912 & 4.548 \\
 & SI-Adv & 100 & 6.060 & 1.784 & 0.977 & 0.144 & 0.576 & 2.081 & 100 & 11.281 & 3.500 & 1.741 & 0.165 & 1.949 & 3.514 \\
 & ITA & 100 & 5.539 & 0.480 & 0.833 & 0.111 & 1.904 & 1.971 & 100 & 9.082 & 1.452 & 1.375 & 0.146 & 3.645 & 2.925 \\
 & Ours & 100 & \textbf{1.801} & \textbf{0.248} & \textbf{0.528} & \textbf{0.105} & \textbf{0.504} & \textbf{0.959} & 100 & \textbf{7.193} & \textbf{1.195} & \textbf{1.237} & \textbf{0.112} & 2.318 & \textbf{2.636} \\ \hline
 \Xhline{1.0pt}
\end{tabular}
}
\vspace{-5mm}
\end{table*}

\firstpara{Adjusting Perturbation Directions with P2S Field}
To ensure that perturbed points remain close to the surface, we use the P2S field \(\mathcal{F}\) to adjust the perturbation directions. Specifically, for the 
$t$-iteration adversarial point \({p'}_i^{(t)}\), we sample the P2S vector \(\mathcal{F}({p'}_i^{(t)})\) and adjust the direction \(\overrightarrow{d_{{p'}_i^{(t)}}}\) as follows:
\begin{equation}
\overrightarrow{d_{{p'}_i^{(t)}}} \  \leftarrow  \   \overrightarrow{d_{{p'}_i^{(t)}}} + \theta ||{p'}_i^{(t)} - p_i||  \cdot \frac{\mathcal{F}({p'}_i^{(t)})}{||\mathcal{F}({p'}_i)^{(t)}||},
\end{equation}
where $\theta$ is a weighting hyperparameter.

\firstpara{Determining Perturbation Magnitudes}
With the refined perturbation directions established, we proceed to determine the perturbation magnitude \(\sigma_{{p'}_i^{(t)}}\) for point \({p'}_i^{(t)}\) following~\cite{liu2022imperceptible}. The $t+1$-iteration adversarial point \({p'}_i^{(t+1)}\) is then obtained as follows:
\begin{equation}
{p'}_i^{(t+1)} = {p'}_i^{(t)} + \sigma_{{p'}_i}^{(t)} \cdot \overrightarrow{d_{{p'}_i^{(t)}}}.
\end{equation}

By iteratively executing the above three steps, our approach creates highly imperceptible adversarial point clouds.

\section{Experimental Results}
\label{sec:experiment}

\subsection{Experimental Setup}

\firstpara{Implementation}
We implement our framework and baseline solutions using PyTorch. The weighting hyperparameter $\theta=0.5$. All experiments are conducted on a workstation with dual 2.40 GHz CPUs, 128 GB of RAM, and eight NVIDIA RTX 3090 GPUs.

\KeyHeadword{Datasets} 
We utilize two public datasets for evaluation: ModelNet40~\cite{wu20153d} and ShapeNet Part~\cite{chang2015shapenet}. Specifically, we randomly sample 1,024 points from each point cloud.

\KeyHeadword{{Victim Models}}  
We select three commonly used DNN classifiers for the attacks: PointNet~\cite{Qi-2017-Pointnet}, DGCNN~\cite{Wang-2019-DGCNN}, and PointConv~\cite{Wu-2019-Pointconv}. These models are trained following the procedures outlined in their respective original papers.

\KeyHeadword{Baseline Attack Methods} 
We select six state-of-the-art techniques as baselines: IFGM~\cite{Dong-2020-FGM_PGD_IMPLEMENT}, PGD~\cite{Dong-2020-FGM_PGD_IMPLEMENT},  SI-Adv~\cite{huang2022shape}, ITA~\cite{liu2022imperceptible}, GeoA$^3$\cite{wen2020geometry} and 3d-Adv\cite{Xiang-2019-3DAdversarialPCD}.

\KeyHeadword{Evaluation Setting and Metrics} 
We configure each attack method to achieve its maximum attack success rate (ASR), defined as the percentage of adversarial point clouds that successfully mislead the victim model.  Under this maximal adversarialness condition~\cite{Tang-2023-ManifoldAttack,tang2024manifoldCons}, we evaluate the imperceptibility of the attacks using six widely recognized metrics: Chamfer distance (CD)\cite{fan2017point}, Hausdorff distance (HD)~\cite{taha2015metrics}, $l_2$-norm ($l_2$), curvature (Curv), geometric regularity (GR)~\cite{wen2020geometry}, and earth mover’s distance (EMD)~\cite{rubner2000earth}. Unless stated otherwise, all results discussed pertain to untargeted attacks.

\begin{figure*}[!t]
\centering
\includegraphics[width=0.88\textwidth]{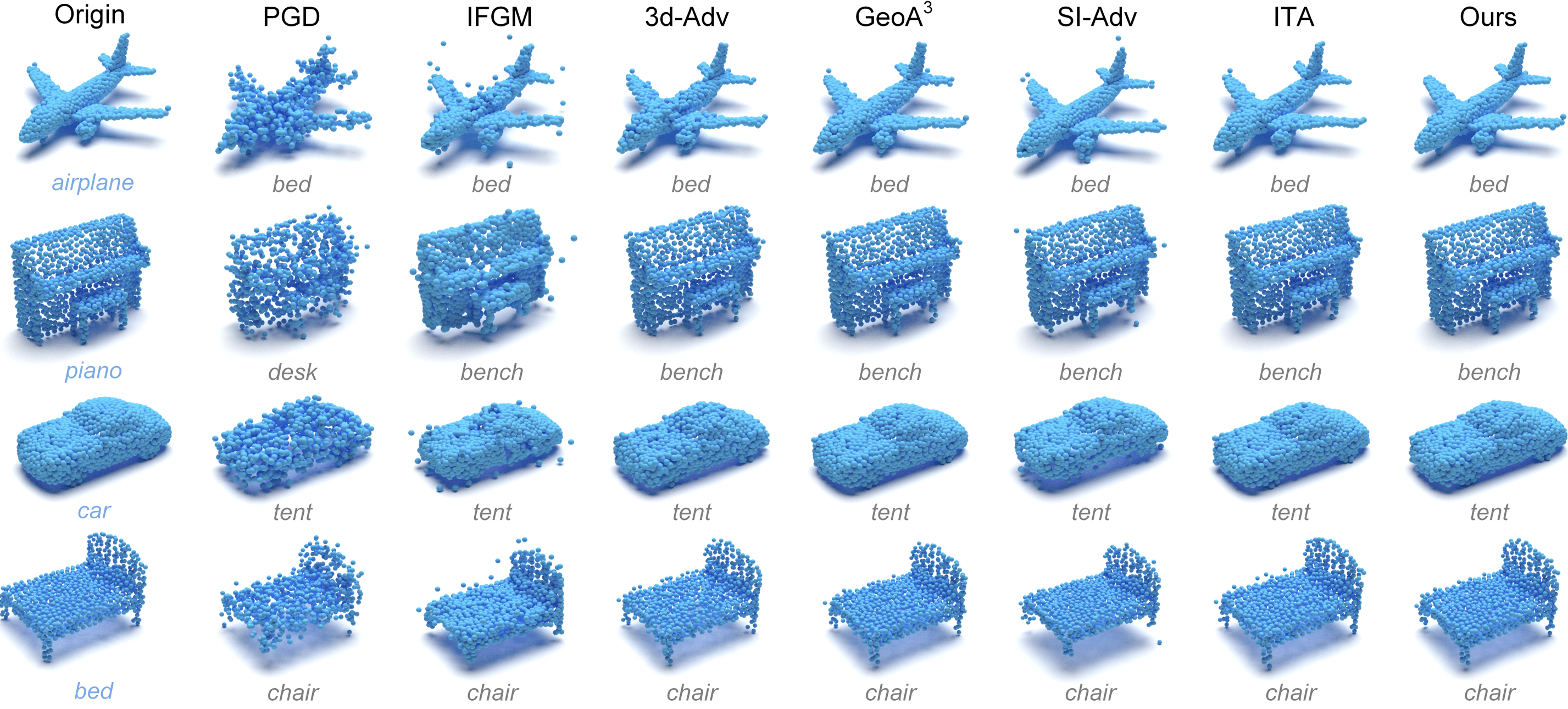}
\vspace{-2mm}
\caption{Visualizations of original and adversarial point clouds generated to fool PointNet on ModelNet40 by various adversarial attack methods. 
The ground truth and predicted labels are marked in blue and gray below the images.
}
\label{fig:VIS_AdvPC}
\vspace{-2mm}
\end{figure*}

\subsection{Performance Comparison and Analysis}

\begin{figure*}[!t]
\centering
\includegraphics[width=0.86\textwidth]{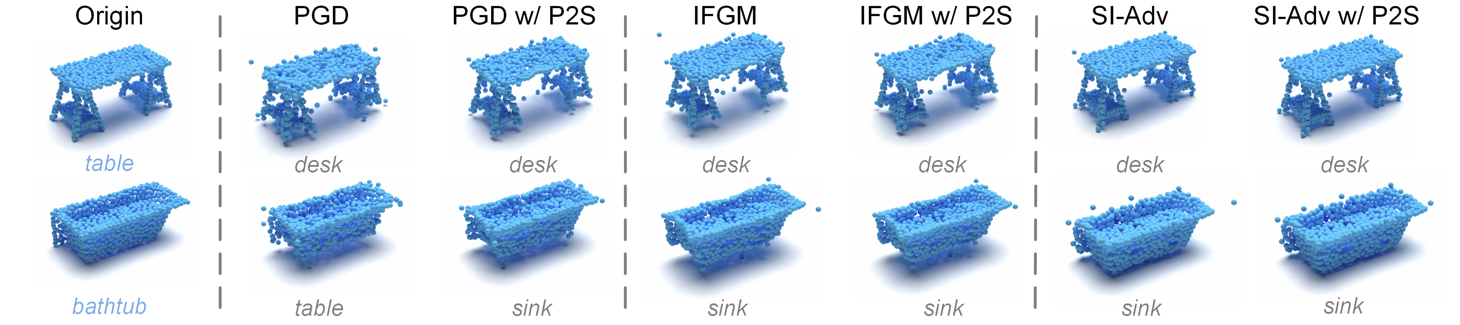}
\vspace{-2mm}
\caption{Visualization of adversarial point clouds generated by various attack methods
in attacking PointNet, with and without 
guidance from the P2S field.
The ground truth and predicted labels are marked in blue and gray below the images.
}
\label{fig:vis2}
\vspace{-3mm}
\end{figure*}

\firstpara{Comparisons with State-of-the-art Methods} 
We evaluate the ASR and imperceptibility of various adversarial attack methods, with the results presented in Tab.~\ref{tab:AttackSummaryTable}. While all methods achieve a 100\% ASR, approaches like PGD exhibit higher distortion, leading to lower performance across most metrics due to their lack of subtlety. In contrast, direction-based methods such as SI-Adv and ITA demonstrate lower distortion, highlighting their effectiveness in maintaining attack imperceptibility. Notably, our P2S field-guided approach surpasses these state-of-the-art methods across the majority of metrics, emphasizing its superior effectiveness in achieving imperceptible adversarial attacks.

\firstpara{Visualization of Adversarial Point Clouds}
We visualize adversarial point clouds generated by various attack methods targeting PointNet on ModelNet40 in Fig.~\ref{fig:VIS_AdvPC}.
Adversarial point clouds from PGD and IFGM show significant outliers, while GeoA$^3$ reduces them with curvature constraints. Directional attacks such as SI-Adv and ITA further minimize visible outliers. Notably, our P2S field-guided solution, which aligns adversarial points with the surface, produces nearly outlier-free point clouds, highlighting the effectiveness and superiority of our method in achieving imperceptibility.


\begin{table}[t!]
\centering
\setlength{\tabcolsep}{1mm}{
\caption{Imperceptibility  of different variants of our solution: without using the P2S field (w/o), using the P2S field in the forward direction (w/ +), and using it in reverse (w/ -).}
\label{tab:Ablation}
\vspace{-1mm}
\scalebox{0.78}{
\begin{tabular}{c|ccccccc}
\hline
\Xhline{1.0pt}
\multicolumn{1}{c|}{\multirow{2}{*}{P2S Field}}  & ASR    & CD  & HD  & $l_2$  & GR  & Curv  & EMD\\
                    & (\%) & ($10^{-4}$) & ($10^{-2}$) &  &  & ($10^{-2}$) & ($10^{-2}$)\\ \hline
    w/o           & 100 & 2.267 & 0.524 & 0.719 & 0.118 & 0.499 & 0.989 \\
    w/ -    & 100 & 2.573 & 0.537 & 0.821 & 0.119 & 0.569 & 1.192 \\
    w/ +    & 100 & \textbf{1.110} & \textbf{0.358} & \textbf{0.366} & \textbf{0.115} & \textbf{0.260} & \textbf{0.449} \\
\hline
\Xhline{1.0pt} 
\end{tabular}}}
\vspace{-2mm}
\end{table}

\firstpara{Ablative Analysis of  P2S Field}
To validate the importance of the P2S field, we compare the results of three configurations: without the field, using the field in reverse, and using the field in its intended forward direction to guide perturbations. The results in Tab.~\ref{tab:Ablation} show that utilizing the P2S field in the forward direction significantly enhances imperceptibility. Conversely, when the field is applied in reverse, causing adversarial points to move further away from the surface, the imperceptibility of the attacks decreases. 
Therefore, we conclude
the critical role of the P2S field in improving the imperceptibility of adversarial attacks.

\begin{table}[t!]
\centering
\setlength{\tabcolsep}{1mm}{
\caption{Comparison of  imperceptibility  of various attack solutions with and without P2S field guidance.}
\label{tab:withP2S}
\vspace{-1mm}
\scalebox{0.78}{
\begin{tabular}{c|ccccccc}
\hline
\Xhline{1.0pt} 
\multicolumn{1}{c|}{\multirow{2}{*}{Attack}}  & ASR    & CD  & HD  & $l_2$  & GR  & Curv  & EMD\\
  & (\%) & ($10^{-4}$) & ($10^{-2}$) &  &  & ($10^{-2}$) & ($10^{-2}$)\\ \hline
PGD w/o P2S        & 100 & 7.155 & 5.025 & 0.981 & 0.302 & 1.624 & 2.315  \\
PGD w/~ P2S     & 100 &\textbf{4.300}  & \textbf{4.402} & \textbf{0.784} & \textbf{0.276} & \textbf{0.740} & \textbf{1.491}  \\ \hline
IFGM  w/o P2S      & 100 & 4.039 & 5.565 & 0.789 & 0.314 & 0.775 & 0.864  \\
IFGM w/~ P2S    & 100 & \textbf{3.366} & \textbf{4.371} & \textbf{0.712} & \textbf{0.275} & \textbf{0.614} & \textbf{0.793}  \\ \hline
SI-Adv  w/o P2S    & 100 & 2.768 & 2.595 & 0.731 & 0.220 & 0.271 &  \textbf{0.725} \\
SI-Adv w/~ P2S  & 100 &  \textbf{2.537} &  \textbf{1.778} &  \textbf{0.644} &  \textbf{0.160} &  \textbf{0.206} &  0.743 \\
\hline
\Xhline{1.0pt} 
\end{tabular}
}}
\vspace{-2mm}
\end{table}

\firstpara{Generalization of  P2S Field} 
To evaluate the generalizability of the P2S field, we integrate it into three established iterative attack methods: PGD~\cite{Dong-2020-FGM_PGD_IMPLEMENT}, IFGM~\cite{Dong-2020-FGM_PGD_IMPLEMENT}, and SI-Adv ~\cite{huang2022shape}. As shown in Tab.~\ref{tab:withP2S}, these methods, when guided by the P2S field, demonstrate significant improvements across most performance metrics under identical parameter settings.
Additionally, we visualize the adversarial samples generated with and without the P2S field in Fig.~\ref{fig:vis2}. 
It is evident that the adversarial point clouds guided by the P2S field exhibit less pronounced outlier points.
These results confirm the broad applicability and effectiveness of the P2S field.

\section{Conclusion}

This paper has introduced a novel point-to-surface (P2S) field-guided framework for imperceptible 3D point cloud attacks. The core idea is to guide perturbed points back to their original underlying surface during attacks. Comprehensive experiments validate the effectiveness of our P2S field-guided attacks in achieving high imperceptibility.

\bibliographystyle{IEEEtran}
\bibliography{GA.bib}

\end{document}